World Scientific
www.worldscientific.com

# Speeding up the Hyperparameter Optimization of Deep Convolutional Neural Networks

Tobias Hinz*, Nicolás Navarro-Guerrero†, Sven Magg‡
and Stefan Wermter§

*Knowledge Technology, Department of Informatics*
*Universität Hamburg*
*Vogt-Kölln-Str. 30, Hamburg 22527, Germany*
*\*hinz@informatik.uni-hamburg.de*
*†navarro@informatik.uni-hamburg.de*
*‡magg@informatik.uni-hamburg.de*
*§wermter@informatik.uni-hamburg.de*

Most learning algorithms require the practitioner to manually set the values of many hyper-parameters before the learning process can begin. However, with modern algorithms, the evaluation of a given hyperparameter setting can take a considerable amount of time and the search space is often very high-dimensional. We suggest using a lower-dimensional representation of the original data to quickly identify promising areas in the hyperparameter space. This information can then be used to initialize the optimization algorithm for the original, higher-dimensional data. We compare this approach with the standard procedure of optimizing the hyperparameters only on the original input.

We perform experiments with various state-of-the-art hyperparameter optimization algorithms such as random search, the tree of parzen estimators (TPEs), sequential model-based algorithm configuration (SMAC), and a genetic algorithm (GA). Our experiments indicate that it is possible to speed up the optimization process by using lower-dimensional data representations at the beginning, while increasing the dimensionality of the input later in the optimization process. This is independent of the underlying optimization procedure, making the approach promising for many existing hyperparameter optimization algorithms.

## 1. Introduction

The performance of many contemporary machine learning algorithms depends crucially on the specific initialization of hyperparameters such as the general architecture, the learning rate, regularization parameters, and many others.[1,2] Indeed, finding an optimal combination of hyperparameters can often make the difference between bad or average results and state-of-the-art performance.[3,4]

*Corresponding author.

Hyperparameter optimization tries to find the optimal set of hyperparameters $\lambda^{(*)}$ which minimizes the generalization error $E$ for the given learning algorithm. This becomes very challenging when the dimensionality of the hyperparameter space increases. Especially, deep neural networks have tens of different hyperparameters that can be adjusted to any given input data set,[3] resulting in a high-dimensional search space. However, hyperparameter optimization problems usually have a low effective dimensionality: even though there is a significant number of hyperparameters, often only a subset of them has a measurable impact on the performance.[5] Yet, for a given learning algorithm, different subsets of hyperparameters matter for different data sets.[5]

We focus on the hyperparameter optimization of one specific learning algorithm, which is widely used: convolutional neural networks (CNNs). One of the biggest challenges is that the evaluation of a given hyperparameter setting for CNNs can take a long time. This is especially the case for deeper models with a potentially high number of filters on each layer. As a result, the inputs to CNNs are often simplified or reduced in size, e.g., by reducing the resolution of images. However, recent studies show that images with higher resolution are advantageous for many classification tasks.[6,7] If hyperparameter values for the same images in different resolutions are similar to each other, we can use this to find appropriate hyperparameters on images with low resolution and then fine-tune them for the same images with high resolution. This is somewhat similar to hyperparameter optimization across data sets,[8,9] where the idea is that hyperparameters that are appropriate for a given data set might be a good starting point for the optimization for similar data sets. However, since different hyperparameters are important for different data sets[5] we do not use hyperparameters from different data sets, but instead identify promising areas in the hyperparameter space on the lower-dimensional representation of the same data.

We apply several hyperparameter optimization algorithms (a genetic algorithm (GA), random search, the tree of parzen estimators (TPEs),[3] and the sequential model-based algorithm configuration (SMAC)[10]) to optimize the hyperparameters of CNNs for image inputs with increasing resolution. That way, we have conceptually the same input data, but in different input dimensions. In the first experiment, we examine the dependencies between the hyperparameters found on the different input sizes to see if there are relationships present. We find that the same hyperparameters are important for a given data set, independent of the image resolution. Furthermore, the optimal value for most hyperparameters also seems to be independent of the image resolution. In the second experiment, we evaluate if this knowledge can be used to speed up the hyperparameter optimization procedure by starting on smaller images and increasing the resolution during the optimization procedure. Figure 1 shows our approach of using increasing image sizes (IIS) to speed up the hyperparameter optimization process.

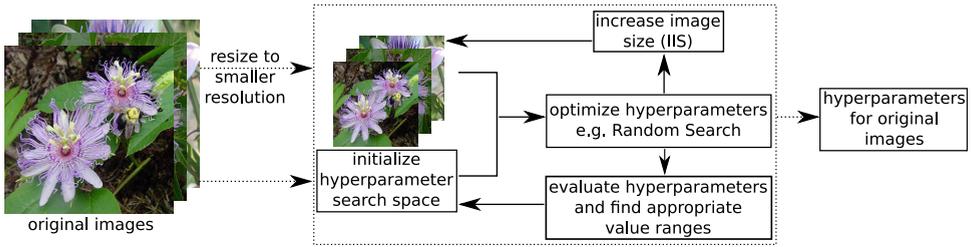

Fig. 1. Overview of our algorithm using IIS: We take the original input images, resize them to a smaller resolution and use a hyperparameter optimization algorithm, e.g., random search or TPE, to find good hyperparameters for these images. Based on the results, we identify good hyperparameter value ranges for all hyperparameters and use them to identify promising areas in the hyperparameter search space. We then increase the image size and run the next iteration of hyperparameter optimization on the larger images, but on the smaller hyperparameter search space initialized from the hyperparameter value ranges identified in the previous iteration. This process can be repeated multiple times until we reach the desired image resolution.

Our experiments suggest that by using increasing image resolutions during the optimization process we can find high-quality hyperparameters in lesser time when compared to the same algorithms optimizing the hyperparameters only on the original sized input images. This method can easily be combined with existing hyperparameter optimization algorithms for CNNs on images to shorten the optimization process, minimizing the amount of computational resources that need to be spent on hyperparameter optimization. For additional information on the experiments and more results see the supplementary material.[11]

## 2. Related Work

Traditionally, the choice of hyperparameters for a given problem is made by the experimenter. However, this requires a significant amount of experience, intuition, and trial and error. Additionally, results are usually not scientifically reproducible and sometimes suboptimal.[12] Recent results indicate that more sophisticated and automated approaches can find better hyperparameters — and thus achieve better results — than humans.[2,3,12–14]

Two of the most widely used methods are also two of the simplest: grid search and random search.[5] In grid search, a pre-determined range of values is chosen for a given set of hyperparameters. Then a grid is constructed through every possible combination of all hyperparameter values. Grid search is easy to implement and trivial to parallelize. However, a big problem is that the grid grows exponentially with the number of hyperparameters. Together with a low effective dimensionality, the grid is likely to be suboptimal since it will potentially cover many spaces of low importance while under-examining hyperparameters in areas of high importance. Random search, on the other hand, draws a random value from a pre-defined distribution for each hyperparameter of interest. It is equally easy to implement and parallelize but can have some advantages in higher-dimensional search spaces. Bergstra and Bengio[5] empirically show that random search performs almost as or equally well in higher-dimensional search spaces while being much quicker than grid search. An extension to random search is introduced by Li *et al.*[15] Their approach, called Hyperband, randomly samples a set of hyperparameter configurations. These configurations are then trained for a certain amount of iterations before they are ranked based on their performance. Then, the best performing configurations are chosen and are trained for an additional number of iterations. This process is repeated until only few configurations are remaining, which are then trained for the maximum number of iterations to find the best configuration.

One of the main problems of optimizing hyperparameters for learning algorithms is that it can take a long time to evaluate a given set of hyperparameters. As a result, Sequential Model-Based Optimization (SMBO) algorithms have been employed in many settings when the performance evaluation of a model is expensive.[3,10,13,16–19] SMBO takes the approach of spending additional computing time on calculating the most promising next hyperparameter instantiation, with the goal that fewer evaluations of the learning algorithm itself are needed. To achieve this, SMBO algorithms employ a probabilistic model to model the black box function $f$, which in this case is the learning algorithm. The model is built with any existing prior knowledge about the problem and point evaluations of $f$.[9] Bayesian optimization[1,4,12,20–24] is one of the most used methods for SMBO, and centers on building a probability model describing the performance given a hyperparameter configuration. The model is then continuously updated with new information gained by sample points providing information about the performance under a given hyperparameter configuration.[1]

Another approach is to employ evolutionary and swarm algorithms for the search of optimal hyperparameters. Population-based optimization methods are well suited for optimization tasks over high-dimensional variable spaces, as they can evaluate many candidate solutions in parallel. They also offer the possibility of combining some sort of random search while utilizing the results of previous evaluations. Miikkulainen *et al.*,[25] Navarro-Guerrero *et al.*,[26,27] Real *et al.*,[28] and Xie *et al.*[29] apply evolutionary algorithms to optimize a subset of hyperparameters for neural networks, while Lorenzo *et al.*[30] use a particle swarm optimization algorithm.

Recently, several approaches use reinforcement learning to find appropriate neural network architectures. Zoph *et al.*[31,32] train a recurrent neural network via reinforcement learning to find neural network architectures that are likely to yield a good performance on specific tasks. Baker *et al.*[33] construct a Q-learning agent that is trained to find CNN architectures that perform well on multiple data sets. Zhong *et al.*[34] also use a Q-learning algorithm to build CNN architectures using individual building blocks. Cai *et al.*[35] introduce an algorithm that transforms existing network architectures which allows to reuse previously trained networks, leading to a large speed-up in the optimization process. However, many reinforcement learning approaches only optimize architectural hyperparameters, while many other hyperparameters such as the learning rate and regularization parameters are manually chosen in the end.

## 3. Methodology

We evaluate our approach in conjunction with several popular hyperparameter optimization algorithms. Similar to Li *et al.*,[15] we use random search, TPE, and, in the final experiment, SMAC to optimize the hyperparameters. Additionally, we also evaluate it in combination with a GA, as they have also shown promising results for the task of hyperparameter optimization.[25,28] Spearmint[12] was excluded since it does not natively support conditional hyperparameters.[36] All optimization algorithms are evaluated with a total of 1500 hyperparameter settings per optimization run.

For all experiments we run the hyperparameter optimization algorithm both on the original images for 1500 evaluations (traditional procedure) and on rescaled images. For the latter, we scale the images to several smaller resolutions and use IIS during the optimization process. Each approach is repeated three times for each data set and optimization procedure. We optimized the following hyperparameters for CNNs: the learning rate, the number of convolutional and fully connected layers, the number of filters per convolutional layer and their size, the number of units per fully connected layer, the batch size, and the L1 and L2 regularization parameters.[11] All other hyperparameters are fixed during the experiments, and similar to Lorenzo *et al.*[30] we stop the training process of a CNN if it does not increase its performance on the validation set for five consecutive epochs. We use a traditional CNN architecture layout, such that each convolutional layer is followed by a max-pooling layer which reduces the input size by a factor of four. Our last convolutional layer is followed by at least one fully connected layer and our final layer is a Softmax layer used for classification.

In the first experiment, we compare the importance of different hyperparameters across different resolutions of the same images in combination with random search, TPE, and the GA. To evaluate this, we need images of sufficient resolution which allows us to scale them down to smaller resolutions in order to get a range of various resolutions for each data set. This excludes popular data sets such as the MNIST and CIFAR data sets, since their resolution is too small ($28 \times 28$ and $32 \times 32$ pixels, respectively). We therefore choose image data sets with a higher resolution (minimum of $96 \times 96$ pixels) and use them to obtain images of the same data sets in various smaller resolutions. In the second experiment, we then test whether using IIS during the optimization process does indeed lead to a speed-up of the optimization process without negative effects on the final network quality.

The first data set is the extended Cohn–Kanade (CK+) data set[37] which consists of images depicting facial expressions of 210 adults, and the task is to classify the displayed emotion. All images were converted to gray scale and resized to four different image sizes of $200 \times 200, 128 \times 128, 64 \times 64$ and $32 \times 32$ pixels, respectively. For the hyperparameter optimization process, we split the data and use 70% for training and the remaining 30% for validation purposes. Since the classes do not have equal amounts of images, we perform the split individually for each class, i.e., of each class we take 70% and add it to the training data, while the remaining 30% are added to the validation data.

The second data set used is the STL-10 data set,[2] which is made up from labeled images acquired from ImageNet. It consists of color images with size $96 \times 96$ pixels and contains 10 classes. Each class has 500 images for training and 800 images for testing purposes. Similarly to the CK+ data set, we converted the images to gray scale and resized them to sizes of $32 \times 32$ and $48 \times 48$ pixels. There exist 10 pre-defined folds containing 100 images from each class for the training set. Training is performed using each of those folds at a time, i.e., using only 1000 images. The reported test set accuracy is then calculated as the average of the accuracy of each of the 10 models on the test set. For the process of hyperparameter optimization, we follow the approach by Swersky *et al.*[4] and use the first fold as training data while using the remaining 4000 images from the training set as our validation set.

In our final experiment, we use the 102 Flowers data set[38] so as not to be biased by the hyperparameters that were found previously. While this data set offers images of very high resolution (minimum $500 \times 500$ pixels), many practitioners rescale the images to a smaller size[39,40] to reduce the number of inputs and the amount of time needed to train the model. Therefore, we reduce the images to $128 \times 128$ pixels as our "maximum" input size, i.e., the input for which the hyperparameters should be optimized are RGB images of size $128 \times 128$ pixels. There are 8189 images in total, and Nilsback and Zisserman[38] provide a pre-defined data split, which gives 2040 images for training and validation, while the remaining 6149 images are used as a test set. The 2040 training and validation images are further split into two equally sized groups, each of which contains 10 images of each flower category. We follow the protocol by Nilsback and Zisserman[38] and use the first 1020 images as a training set to optimize the hyperparameters, evaluating their performance on the other 1020 images.

## 4. Importance of Hyperparameters

We will now examine the importance of the different hyperparameters for different resolutions and the similarity of hyperparameter values across different resolutions of the same images. All three optimization algorithms (random search, TPE, GA) found similar values for the different hyperparameters for all resolutions. Especially, the chosen learning rate, the batch size, the L1 and L2 regularization penalties, and the general architecture (i.e., number of layers) are very similar for all optimization algorithms. Minor differences can be found in the number of filters and units per convolutional or fully connected layer. The hyperparameters found by the GA and TPE typically performed better than those found by random search.[11]

To evaluate the importance of the various hyperparameters, we use a variant of analysis of variance (ANOVA), called functional ANOVA, to analyze the importance of different subsets of hyperparameters as suggested by Hutter *et al.*[14] We define the importance of a subset of hyperparameters as the amount of variance it accounts for in the validation error. The higher the amount of variance in the validation error that is explained by a certain subset of hyperparameters, the higher its importance. If the explained variance is high this means it is important to find "good" values for this subset of hyperparameters, since a "bad" value will likely lead to suboptimal results. Conversely, hyperparameters that have little impact on the variance can be neglected in the optimization process, since they have little impact on the final performance.

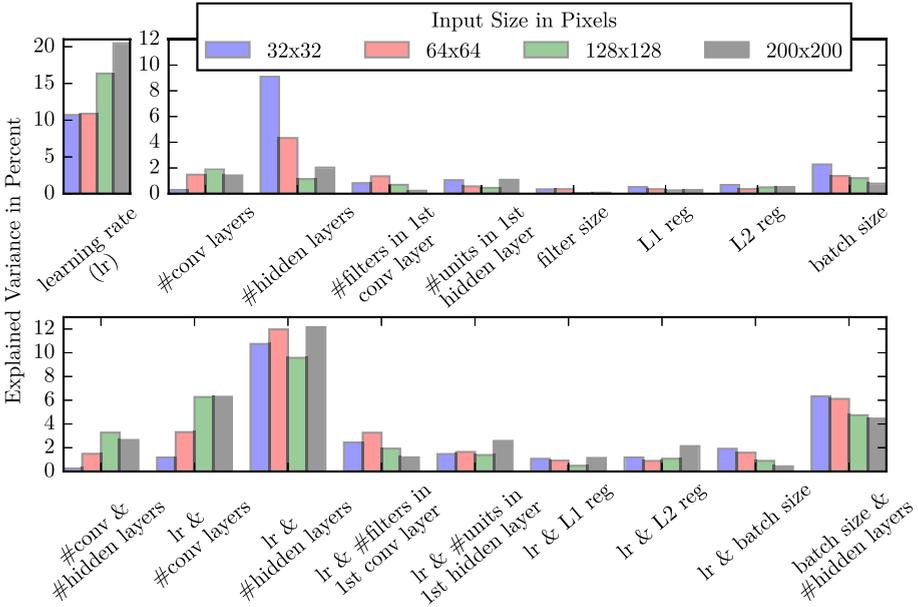

Fig. 2.   Explained variance of the validation error in percent on the CK+ data set. Depicted are the effects of the 18 most important hyperparameter subsets, aggregated from the data of all runs of the GA, random search, and TPE.

Figure 2 shows an overview of the — on average — most important hyperparameter subsets of the CK+ data set and their impact on the validation error. We can observe that the importance of hyperparameter subsets stays somewhat constant across the different input sizes. The only subsets that deviate strongly from this are the subsets that include the number of convolutional and fully connected layers. However, this is to be expected as the number of convolutional layers is directly dependent on the input dimension (since we insert a max-pooling layer after each convolutional layer) and the number of hidden layers might be dependent on the number of convolutional layers. Other hyperparameters, such as the batch size and regularization parameters, are similarly important for all input sizes. Indeed, the top five and top 10 of the most important hyperparameter subsets are virtually identical for all input sizes.

Figure 3 shows the impact of the learning rate on its own on the average validation error of the CK+ data set. We can see that the learning rate's impact is very similar across all input settings. Figure 4 depicts the impact of the combination of the learning rate and the number of hidden layers on the validation error of the CK+ data set. Again, this is very similar across all input sizes, identifying a learning rate between 0.1 and 0.01 together with one hidden layer as good parameters. Both Figs. 3 and 4 further indicate that not only the importance of hyperparameters is the same for different input sizes, but even the general "best" value for a given hyperparameter seems to be closely correlated. These results are closely reflected on the STL-10 data set, see the supplementary material.[11]

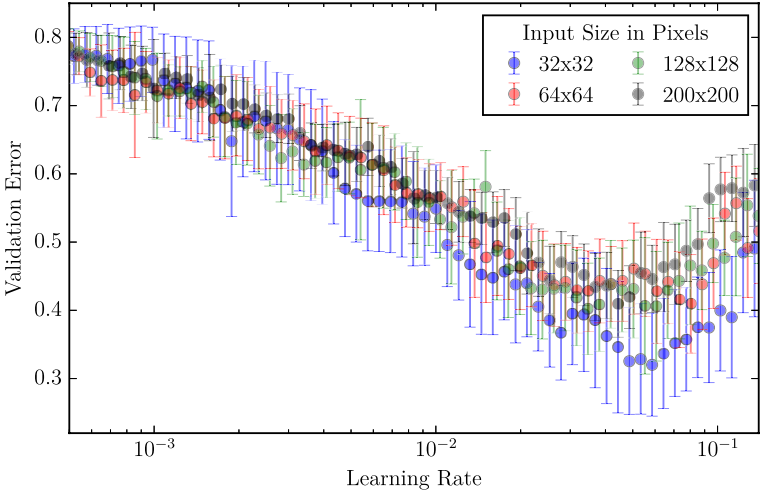

Fig. 3. Relationship between the learning rate and the validation error (CK+ data set), aggregating the data of all three algorithms on the respective image resolutions.

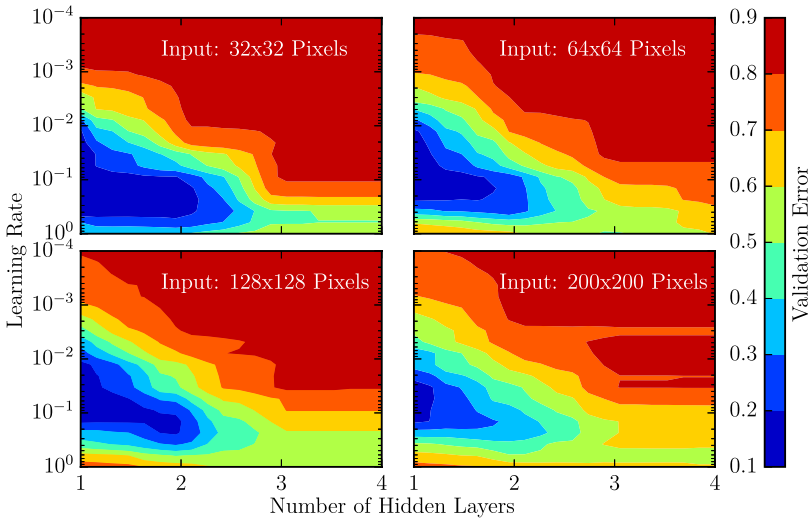

Fig. 4. Relationship between the learning rate, the number of hidden layers, and their impact on the validation error for the CK+ data set, aggregating the data of all three algorithms on the respective image sizes.

The experiment also showed that the different hyperparameters' importance is approximately consistent across the various input sizes. The only exception to that are hyperparameters that include the number of layers. Due to the max-pooling layers, CNNs with smaller inputs cannot have as many convolutional layers as CNNs with bigger inputs. Additionally, for instance, for an input size of $32 \times 32$ pixels, a CNN might still perform reasonably well with only one convolutional layer, even if two layers lead to an improvement of performance. For an input size of $96 \times 96$ pixels, or even $200 \times 200$ pixels, on the other hand, a CNN with only one convolutional layer does not perform well at all. This is an inherent problem of the optimization process across different input dimensions and most likely means that the optimal number of convolutional layers has to be found for each specific input size. A good starting point for the number of filters per convolutional layer can be inferred from smaller input dimensions, at least for convolutional layers that are present in CNNs for smaller inputs.

For a more unbiased evaluation of the hyperparameters, we also test some hyperparameter settings on held-out test sets. For these tests, we choose the hyperparameter settings of each algorithm that performed best on the respective validation set. We adhere to the common guidelines about the test sets as detailed by Khorrami *et al.*[41] for the CK+ data set and Coates *et al.*[2] for the STL-10 data set.

Both the GA and the TPE algorithm significantly outperform random search[11] when no additional regularization methods are applied. All in all, the results are similar to the results obtained on the validation sets, which indicates that the hyperparameters are not fit specifically to the validation set, but rather are appropriate hyperparameters for this kind of input. We also find that the accuracy usually increases with a higher image resolution, highlighting the importance of a high enough resolution for optimal performance.[6,7]

## 5. Using IIS for Hyperparameter Optimization

The previous experiment suggests that hyperparameters are approximately of the same importance, independent of the image resolution, and that good hyperparameter values are similar across image resolutions. The traditional approach is to take the data as is and then run an algorithm to optimize the hyperparameters for the given model. Our approach, in contrast, does also make use of the same images, but in smaller resolutions. Due to this, we rescale the images of the 102 Flowers data set to sizes of $64 \times 64$ and $32 \times 32$ pixels.

Our pipeline for the optimization process is then as illustrated in Sec. 1, Fig. 1: we use our algorithms to optimize the hyperparameters of a CNN that receive as input images of size $32 \times 32$ pixels for 750 evaluations. The hyperparameters obtained through this are then used to initialize the algorithms for the optimization process on the images of size $64 \times 64$ pixels for another 500 evaluations. Finally, these hyperparameters are used to initialize the algorithms to optimize the hyperparameters for images of size $128 \times 128$ pixels for the final 250 evaluations. With this strategy, we expect to need less time to arrive at hyperparameters that are comparable in performance to those obtained by the same algorithms run on the images of resolution $128 \times 128$ pixels for all 1500 evaluations. In addition to the previously used hyperparameter optimization algorithms, we now also use the SMAC[10] as an additional state-of-the-art optimization procedure for further validation.

We will now have a look at how the different optimization algorithms perform during the two strategies. For more details on the initialization of the algorithms and the obtained hyperparameters and results on the test, set see the supplementary material.[11] Figure 5 shows how the best validation error developed during the different optimization processes, averaged over three runs of each optimization process for each strategy. The left column shows the development of the validation error per 50 evaluations for all algorithms, while the right column depicts the same information in relation to the elapsed time. Table 1 shows the exact amount of time in minutes it took to perform each of the optimization procedures, averaged over three runs for each method.

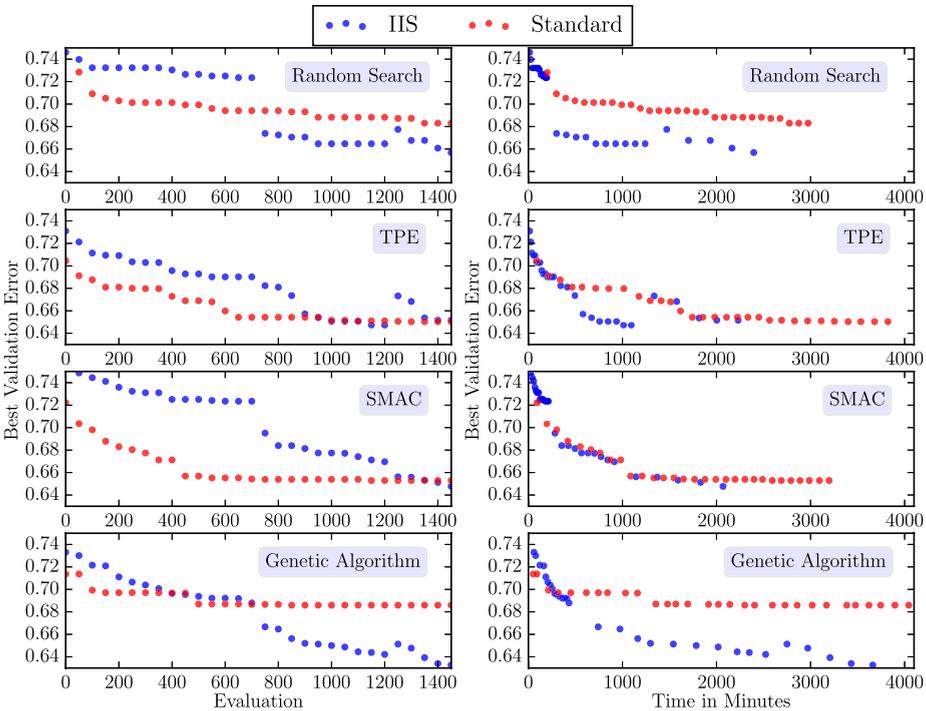

Fig. 5. Best average validation error for all algorithms on the 102 Flowers data set. The progress is visualized over the number of evaluations (left column) and the amount of elapsed time in minutes (right column).

Table 1.  Amount of time in minutes it took for each of the optimization methods shown in Fig. 5, and the time reduction of using IIS with respect to the standard approach.

| | Random Search | TPE | SMAC | Genetic Algorithm |
|---|---|---|---|---|
| Standard | $2974 \pm 77$ | $3822 \pm 294$ | $3197 \pm 885$ | $4043 \pm 199$ |
| Increasing image size | $2396 \pm 335$ | $2230 \pm 46$ | $2069 \pm 103$ | $3662 \pm 386$ |
| Time reduction | 19% | 42% | 35% | 9% |

For the standard procedure, when the optimization is only performed on the full-sized images, we see that, except for random search, the algorithms find their best solution after 500–750 evaluations. For optimization over IIS, we see that the algorithms improve their solution continuously until the final evaluation. The solution usually improves between the 750th and 800th evaluations, when the image size is increase to $64 \times 64$ pixels. However, when the image size is increased to $128 \times 128$ pixels (evaluation 1250), we often observe a drop in performance and it takes a few evaluations to improve upon the previous best performance.

When we look at the development of the validation error over time (right column of Fig. 5), we see that the procedure with IIS needs between 9% and 42% less time than the standard approach. The difference in time is especially big for the TPE and SMAC algorithms (see Table 1), while it is less pronounced for the GA and random search. For both the GA and random search, the final solutions are also significantly better when they are found while using increasing image sizes as opposed to when the optimization process is performed only on the originally sized images.

The increase in performance with random search and the GA is likely due to the fact that we reduce the size of the hyperparameter search space to "good" value ranges for the different hyperparameters whenever we increase the image size. The same effect could probably be achieved by decreasing the search space of the hyperparameters during the traditional optimization process after a given number of evaluations. However, as we showed in Sec. 4, it is possible to find good hyperparameter value ranges with smaller-scaled images, leading to a speed-up in the training of the CNNs. Summing up, the approach using IIS never leads to worse performance, but leads to a significant reduction of the time needed for the optimization process (especially for TPE and SMAC), and sometimes also leads to significantly better results (especially for the GA and random search).

In our second experiment, we could illustrate the power of using smaller input sizes to find promising areas in the hyperparameter space, before optimizing the hyperparameters on the final input size. With this approach, multiple optimization algorithms were able to find comparative results in less time than when they were run on the full-sized images. This is a very promising result since this approach is independent of the underlying optimization algorithm. It can instead be applied to any optimization methodology for which it is possible to reduce the dimensionality of the data in a meaningful way.

To make sure that the technique of using IIS during the optimization process does not negatively affect the final CNN performance, we also compare the obtained hyperparameters on the 102 Flowers test set of 6149 images. The test set is not balanced, with different classes containing between 20 and more than 200 examples. Following the standard approach,[38] we report the average classification error averaged over all classes. All settings perform similarly well,[11] i.e., around 46% accuracy on the test set, when no additional regularization methods are applied, with random search being 1% less accurate than the more sophisticated algorithms. This indicates further that the success of using IIS during the optimization process is independent of the underlying optimization algorithm since the final performance of all hyperparameter settings is very similar.

## 6. Conclusion

In this work, we presented an approach to reduce the time needed for hyperparameter optimization of deep CNNs. One of the main challenges of optimizing hyperparameters in CNNs is that the training process can take a very long time, which makes it expensive to evaluate many different hyperparameter combinations. To deal with this, we propose to first find promising hyperparameter values on a lower-dimensional representation of the original data. To test this, we rescale images to a lower resolution and optimize the hyperparameters of CNNs on those smaller images. The results of this are then used to initialize the hyperparameter space of the optimization process on the original-sized images. In theory, this process can be repeated several times, i.e., the hyperparameters can be optimized on multiple smaller representations that increase in size until we reach the original data size. Our first experiment in Sec. 4 shows that hyperparameter importance and hyperparameter values are mostly consistent across different resolutions of the same images. Our second experiment in Sec. 5 uses this knowledge and shows that using IIS can speed up the optimization process significantly. Furthermore, identifying important hyperparameters and good value ranges early on in the optimization process can also help to achieve a higher accuracy on the original images after the optimization process is completed.

We investigate this approach on a more theoretical level in the first experiment on two different data sets. Here, we optimize the hyperparameters independently on the same images but with different resolutions. We observe that a significant number of hyperparameters overlap in their values, independent of the image resolution. Additionally, we find that the importance of different hyperparameter subsets in relation to each other stays roughly the same across different resolutions. This can be used to quickly identify important hyperparameters on lower-dimensional data.

In the second experiment, we test this on a third data set and show that an approach using this methodology finds good hyperparameters faster than the traditional approach of optimizing the hyperparameters only on the original image size. This concept is generally applicable to many different hyperparameter optimization methodologies and is not restricted to a specific methodology. We applied the procedure on four hyperparameter optimization algorithms (random search, TPE,[3] SMAC,[10] and a GA[11]) and found it to work well on each one of them. Furthermore, this technique is easily extensible with other methods for speeding up the hyperparameter optimization process, such as parallelization, extrapolating learning curves,[42] or algorithms developed for speeding up the hyperparameter optimization procedure such as Fabolas[24] and Hyperband.[15] Moreover, this might not only be applicable to images, but to any input whose dimensions can be reduced in a meaningful way, e.g., through common dimensionality reduction algorithms like PCA. However, while we show that it works well on CNNs in conjunction with image classification, future work needs to test if this is also the case for other tasks and other ways of dimensionality reduction.

## Acknowledgments

The authors gratefully acknowledge partial support from the German Research Foundation DFG under Project CML (TRR 169) and the European Union under Project SECURE (No. 642667).